\documentclass[journal]{IEEEtran}
\usepackage{graphicx}
\usepackage{xcolor}
\usepackage{amsmath}
\usepackage{amssymb}
\usepackage{multirow}
\usepackage{array}
\usepackage[hidelinks]{hyperref}

\usepackage{algorithm}
\usepackage{algorithmic}
\usepackage{balance}
\usepackage{tikz}
\usetikzlibrary{shapes.geometric,arrows.meta,calc}

\IEEEoverridecommandlockouts

\title{\LARGE \bf{\textbf{Self-Sensing Compliant Anthropomorphic Hand for Speed-Dictated Shared Force Control in Teleoperation}}}

\author{\parbox{\textwidth}{\centering Huijiang~Wang$^{1}$, Yu~Zhang$^{1}$, Rui~Chen$^{1,2}$, Tristan~Bonato$^{1}$, Jaemin~Eom$^{1}$, Kai~Junge$^{1}$,\\ Cheng~Pan$^{1}$, Benhui~Dai$^{1}$, Yue~Xie$^{3}$, Josie~Hughes$^{1}$}%
\thanks{$^{1}$CREATE Lab, \'{E}cole Polytechnique F\'{e}d\'{e}rale de Lausanne (EPFL), Lausanne, Switzerland. (Email: huijiang.wang@epfl.ch)}%
\thanks{$^{2}$Institute of Mechanical Intelligence, School of Advanced Studies Sant'Anna (SSSA), Pisa, Italy.}%
\thanks{$^{3}$Department of Computer Science, Loughborough University, Loughborough, United Kingdom.}%
\thanks{Project website:
    \url{https://huijiang-wang.github.io/selfsensing-hand/}}
}

\begin{document}

\maketitle
\thispagestyle{empty}
\pagestyle{empty}

\begin{abstract}
In vision-based teleoperation of robotic hands, the operator receives no haptic feedback and therefore struggles to regulate the grasp force, while adding tactile sensors to the fingers increases wiring, cost and integration complexity. Compliant hands offer an alternative, as their mechanical compliance, usually treated only as a passive property for safe and adaptive grasping, can also serve as a source of information. Here, we show that the elastic transmission of a compliant, tendon-driven anthropomorphic hand can itself act as the sensor for shared-control telemanipulation. Series springs in the tendon routing are read by potentiometers in a forearm-mounted motor box, from which the hand estimates its joint states and finger forces without any external sensors. Building on this self-sensing, we develop a contact-aware shared-control framework for VR-based teleoperation. The robot hand is controlled through a mapping from the operator's hand kinematics until a contact is detected proprioceptively. Control then switches to a finger-level force controller whose set point is determined by the operator's grasping speed, and control authority can be allocated to individual fingers. Experiments validate the system's force estimation and increased grasp success rate compared with direct teleoperation for fragile and heavy objects. These results show that compliance can serve not only for passive adaptation but also as an intrinsic sensing resource for physical human--robot collaboration.
\end{abstract}


\section{Introduction}

Teleoperation allows human operators to guide robots through tasks in which full autonomy remains unreliable, and it has become the main route for collecting human demonstrations for robot learning \cite{zhang2018deep,qin2023anyteleop}. For multifingered anthropomorphic hands, the operator's hand is commonly tracked with RGB-D cameras \cite{handa2020dexpilot,gao2022efficient}, data gloves \cite{shaw2024bimanual} or, increasingly, virtual-reality (VR) headsets that track the bare hand without tethered hardware \cite{cheng2024open,ding2024bunny,junge2025adapt}. These interfaces map human finger motion to the robot accurately in free space, but they are purely kinematic because the operator sees the scene but cannot feel it. Once the fingers touch an object, the operator has no direct means to regulate the grasp force, so that fragile objects are over-compressed and heavy objects might slip. This problem is further aggravated by self-occlusion of the fingers and by the kinematic mismatch between the human and the robotic hand \cite{li2020mobile,li2021survey}. Improving such teleoperation therefore calls for ways to regulate the grasp that do not rely on complex haptic interfaces.

Shared control addresses this limitation by distributing control authority between the human and the robot, combining human reasoning and decision-making with the sensing and control capabilities of the robot \cite{li2023classification,huang2025human}. At the hand level, the operator provides coarse finger commands while the robot hand regulates the details of the contact, such as grasp force or finger stiffness \cite{michelman1994shared,griffin2003feedback}, which improves grasp success and reduces operator effort \cite{zhu2023shared,laghi2022target}. A prerequisite of every such strategy is that the robot estimates its contact state reliably, both to know when to take over from the operator and to regulate the force once it has.

\begin{figure}[t]
    \centering
    \includegraphics[width=1\linewidth]{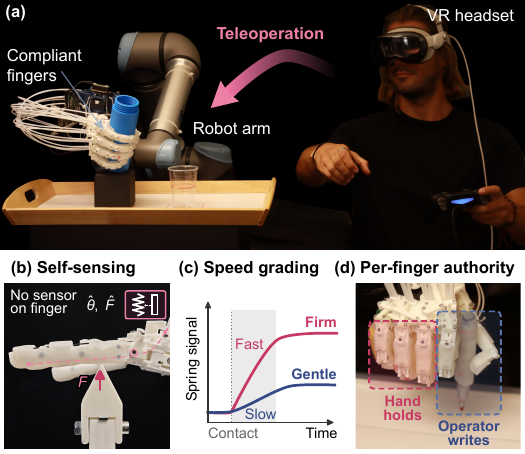}
    \caption{Speed-dictated shared control with a self-sensing compliant hand. (a) VR teleoperation of a compliant hand on a robot arm. (b) Joint angles and forces estimated without sensors on the fingers. (c) Speed-graded grasping: the closing speed selects a gentle or a firm grasp. (d) Per-finger allocation of control authority.}
    \label{fig:framework}
\end{figure}

This contact information is usually obtained from tactile sensors integrated into the fingers, such as six-axis force/torque sensors in the fingertip \cite{kim2020six,mao2025ultralight}, soft magnetoelectric fingers \cite{xu2024soft}, and tactile skins on the phalanges \cite{li2024tactile,almeida2025role}. Integrating them into a multifingered hand, however, requires wiring through the joints, fitting the sensors into the confined form factor of the finger, and achieving the necessary sensitivity \cite{mao2025ultralight}. Contact forces can also be inferred from images of a compliant hand \cite{zhu2025forces}, but vision degrades under the occlusions that occur during grasping.

An alternative to tactile sensing is to infer the contact from proprioception, i.e., from the internal signals of the actuation. Compliant hands are particularly suited to this approach. Underactuated, tendon-driven designs with elastic elements conform passively to objects and tolerate imperfect commands \cite{della2018toward,li2024tactile,junge2025spatially}, which makes them well suited for teleoperation \cite{junge2025adapt}. Their compliant transmission deforms with finger-object interaction, and this deformation encodes the external load, much as Golgi tendon organs, placed in series between muscle and tendon, provide humans with a sense of muscle tension \cite{jami1992golgi}. In tendon-driven hands, this principle has been exploited through remote contact-force sensors \cite{kim2023multiparameter}, fiber-optic tendons \cite{yi2023underactuated}, and series elastic actuators that estimate the grasp posture and regulate the grasp force \cite{lee2025tendon}. In teleoperated hands, the tendon tension has been used for low-level finger force control \cite{bridgwater2012robonaut} or rendered to the operator through a haptic glove \cite{yi2026multifingered}. In hand-level shared control, the contact typically switches the hand to a fixed grasp behaviour. It remains underexplored how the operator can set the force requirement of a grasp fluently and without any additional device.

In this work, the operator sets the grasp force through the speed of a contact-rich gesture, namely the speed with which they close the hand on the object. Humans adapt the kinematics of reaching and grasping to the object and the task, for instance with a longer deceleration phase when grasping a fragile object \cite{marteniuk1987constraints}. The closing speed is treated as an implicit command that the operator already produces. When deployed on a compliant cable-driven hand, this speed does not need to be tracked by an additional external device, as it can be decoded from the rate at which the series spring is loaded once the finger-object interaction happens.

We realize this on a tendon-driven anthropomorphic hand \cite{junge2025spatially}, into whose forearm-mounted motor box we integrated a series-elastic sensing unit for each finger and for the thumb (Fig.~\ref{fig:framework}). Springs in the metacarpophalangeal (MCP) tendons of the four fingers and in the two carpometacarpal (CMC) tendons of the thumb are read by linear potentiometers, so that all sensing electronics remain outside the hand. From these spring signals and the motor states, the hand estimates its joint configuration and contact forces. On this basis, we build a speed-dictated shared-control framework for VR-based teleoperation. The operator controls the hand freely until a proprioceptively detected contact transfers grasp regulation to the hand, the loading rate at contact selects a gentle or a firm grasp, and control authority can be allocated to individual fingers. The main contributions of this paper are:
\begin{enumerate}
    \item A self-sensing approach that estimates joint states and fingertip forces from the series springs of a compliant tendon-driven hand.
    \item A contact-aware shared-control framework in which a proprioceptively detected contact transfers grasp regulation to the hand and the operator's approach speed grades the grasp force.
    \item A finger-level allocation of control authority that lets the hand hold one object while the operator manipulates another.
    \item Validations showing that speed-dictated grasping and per-finger control authority enable, without haptic feedback, the reliable handling of both fragile and heavy objects and concurrent grasping and tool use.
\end{enumerate}

\section{Materials and Methods}
\label{sec:platform}
\label{sec:methods}

We first describe the hand and its series-elastic sensing (Sections~\ref{sec:handDesign} and \ref{sec:selfSensing}), then the teleoperation interface (Section~\ref{sec:teleop}) and the speed-dictated shared control built on it (Section~\ref{sec:sharedControl}). Implementation details and experimental protocols are given in Section~\ref{sec:setup}.

\subsection{Hand and Series-Elastic Sensing}
\label{sec:handDesign}
\label{sec:springSensing}

\begin{figure}[t]
    \centering
    \includegraphics[width=\linewidth]{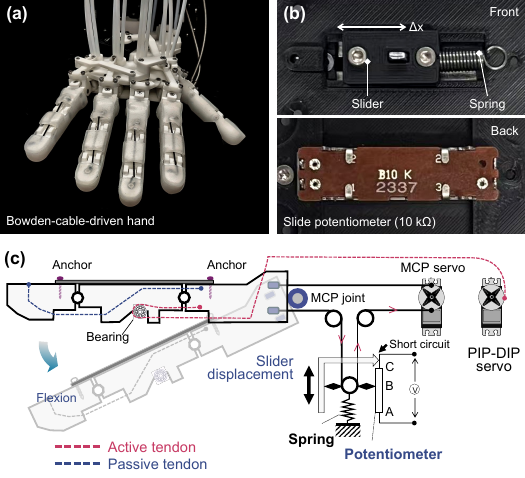}
    \caption{Compliant hand and spring sensing unit. (a) Bowden-cable-driven hand. (b) Sensing unit: the spring elongation $\Delta x$ moves the slider of a linear potentiometer. (c) Finger tendon routing, with the spring and potentiometer in series with the MCP agonist tendon.}
    \label{fig:handPotentiometer}
\end{figure}

The hand (Fig.~\ref{fig:handPotentiometer}a) follows the Bowden-cable architecture of the ADAPT Hand V2.1 \cite{junge2025spatially}, with fingers whose proximal and distal interphalangeal (PIP and DIP) joints are coupled \cite{zhang2026modular}. It has 21 degrees of freedom actuated by 15 servomotors in a forearm-mounted motor box above the wrist: two per finger, driving the MCP joint and the coupled PIP--DIP joints, four for the thumb, driving its CMC1 (abduction--adduction), CMC2 (flexion--extension), MCP and interphalangeal (IP) joints, one that spreads the four fingers through an abduction synergy, and two for the wrist pitch and yaw. The fingers therefore carry no electronics or wiring.

To make the hand self-sensing, we integrated a series-elastic sensing unit into its transmission. The MCP joint of each finger is driven through a dual-pulley antagonistic mechanism, and a tension spring is inserted in series with the agonist tendon (Fig.~\ref{fig:handPotentiometer}c). The spring serves two purposes: it adds compliance that absorbs impacts and lets the finger conform to objects, and it converts the tendon tension into a displacement. This displacement moves the slider of a linear potentiometer (Fig.~\ref{fig:handPotentiometer}b), whose reading, the spring signal $s$ in ADC counts, is proportional to the spring elongation and thus to the tendon tension. The spring stiffness sets the trade-off between force range and sensitivity. The two thumb CMC tendons carry the same unit, which gives six sensing channels: the index, middle, ring and little-finger MCP and the thumb CMC1 and CMC2.

\subsection{Self-Sensing of Joint States and Contact Force}
\label{sec:selfSensing}
\label{sec:sensingPrinciple}

\begin{figure}[t]
    \centering
    \includegraphics[width=\linewidth]{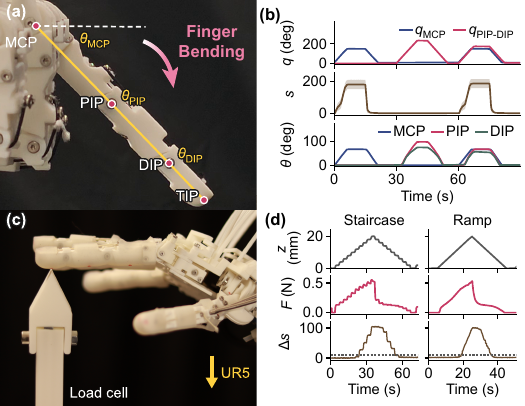}
    \caption{Self-sensing calibration. (a) Optical ground truth of the joint angles. (b) Free-motion training data. (c) Force calibration with the UR5 and a load cell. (d) Staircase and ramp loading.}
    \label{fig:selfSensingMethod}
\end{figure}

Each spring signal combines a posture and a contact component,
\begin{equation}
s = s_0(q) + \Delta s, \qquad \Delta s = h(F, \mathbf{x}_c),
\label{eq:springDecomposition}
\end{equation}
where $s_0(q)$ is the free-motion signal at the motor positions $q$, which rises with flexion of the joint that loads the spring because of elastic and frictional forces in the tendon routing (Fig.~\ref{fig:selfSensingMethod}b), and $\Delta s$ is the change caused by a contact force $F$ at location $\mathbf{x}_c$. Because $\Delta s$ scales with the torque $F d_c$ about the MCP joint, where $d_c$ is the distance of the contact from the joint, the spring responds to contacts on any phalanx, with a sensitivity that decreases towards the palm. Static friction in the Bowden tubes and pulleys adds a dead zone at contact onset and hysteresis between loading and unloading.

\paragraph{Joint-state estimation}
\label{sec:jointEstimation}
Because of dead zones in the tendon routing, the joint angles are nonlinear functions of the motor positions, and the posture component $s_0(q)$ adds information on the MCP angle. Each joint angle $j \in \{\text{MCP}, \text{PIP}, \text{DIP}\}$ is therefore estimated from the standardized input $\mathbf{x} = [q_{\text{MCP}}, q_{\text{PIP--DIP}}, s]^\top$ by a support vector regression (SVR) with a Gaussian kernel,
\begin{equation}
\hat{\theta}_j(\mathbf{x}) = \sum_{i=1}^{N} \beta_{j,i}\, \exp\!\left(-\gamma \left\| \mathbf{x} - \mathbf{x}_i \right\|^2\right) + b_j,
\label{eq:svr}
\end{equation}
where $\mathbf{x}_i$ are the training inputs, $\gamma$ is the kernel width, and $\beta_{j,i}$ and $b_j$ are fitted with an $\epsilon$-insensitive loss. The thumb CMC joints are estimated in the same way from their own motor position and spring signal, and the thumb MCP and IP joints from their motor positions alone.

\paragraph{Contact-force estimation}
\label{sec:forceEstimation}
Because of the hysteresis caused by static friction, the relation between force and spring signal differs between loading, i.e., while the contact force increases, and unloading. We model only loading, which is the phase relevant for detecting and regulating contact, by a linear model with an activation threshold,
\begin{equation}
\hat{F} = F_{\text{th}} + a\,\Delta s ,
\label{eq:linearForce}
\end{equation}
where $\Delta s$ is taken relative to the unloaded level before contact, $F_{\text{th}}$ is the force required to overcome static friction before the spring engages, and $a$ is the channel gain. The model applies once the spring has engaged, $\Delta s > 10$\,counts (more than ten times the resting noise); below this level, the force is only known to lie between zero and $F_{\text{th}}$. Since $\Delta s$ scales with $F d_c$, both parameters depend on the loaded phalanx. Both models are fitted per channel on calibration data (Section~\ref{sec:setup}).

\subsection{Teleoperation Interface}
\label{sec:teleop}

\begin{figure}[t]
    \centering
    \includegraphics[width=\linewidth]{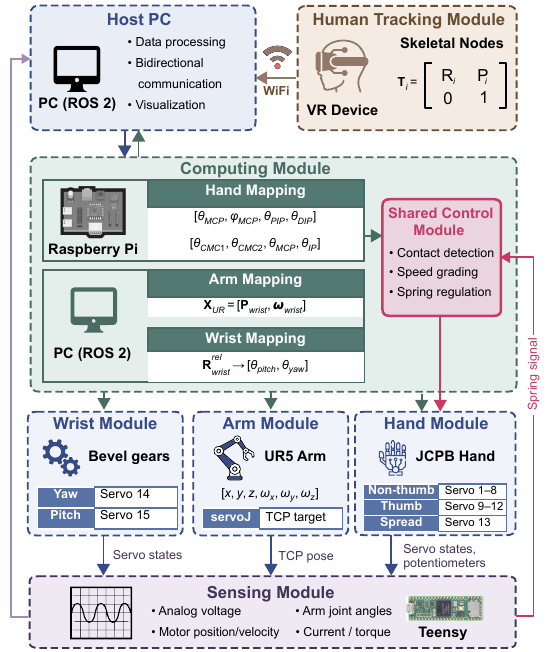}
    \caption{Teleoperation framework. Hand poses tracked by the VR headset are mapped to commands for the hand, arm and wrist; the spring signals feed the shared-control module (Section~\ref{sec:sharedControl}).}
    \label{fig:teleopFramework}
\end{figure}

The operator's right hand is tracked by an Apple Vision Pro (AVP) headset, and the operator receives only visual feedback (Fig.~\ref{fig:teleopFramework}; Video~S1). The joint angles of the operator are obtained from the relative rotations of adjacent nodes of the 27-node AVP hand skeleton \cite{park2024avp}. The MCP angle is mapped one-to-one to the robot, and since the robot PIP and DIP joints are coupled, the sum of the human PIP and DIP angles is mapped linearly onto the range of the PIP--DIP motor; the thumb is mapped analogously. The wrist pose commands the tool pose of a UR5 arm, and the wrist rotation relative to the palm is projected onto the pitch and yaw of the robot wrist. Under contact, the series spring stretches, which is the signal used by the shared control. Without the shared-control layer, the system reduces to direct kinematic teleoperation, which serves as the baseline in Section~\ref{sec:experiments}.

\subsection{Speed-Dictated Shared Control}
\label{sec:sharedControl}

\begin{figure}[t]
    \centering
    \includegraphics[width=\linewidth]{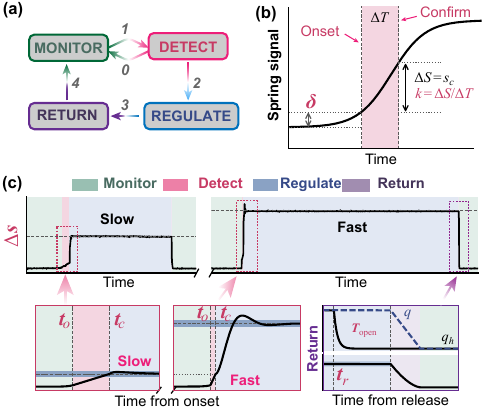}
    \caption{Speed-dictated shared control. (a) Per-finger state machine. (b) Contact detection and loading rate. (c) Spring change of the index finger during a slow and a fast closure, with the controller state, and the delayed return of control authority after release.}
    \label{fig:sharedControl}
\end{figure}

The shared-control layer sits between the motion mapping, which produces the operator command $q_h$, and the motor commands $q$. For each sensed finger, it (1)~detects the contact, (2)~measures how fast the operator closed the finger, and (3)~regulates the contact at a force level selected by this speed (Fig.~\ref{fig:sharedControl}).

\begin{table}[t]
\centering
\caption{State Transitions of the Per-Finger Controller}
\label{tab:fsm}
\setlength{\tabcolsep}{4pt}
\footnotesize
\begin{tabular}{@{}l>{\raggedright\arraybackslash}p{4.6cm}@{}}
\hline
Transition & Condition \\
\hline
\textsc{Monitor} $\rightarrow$ \textsc{Detect} & onset: $s - \bar{s} > \delta$; store $t_o$, $s_o$ \\
\textsc{Detect} $\rightarrow$ \textsc{Monitor} & noise: $s - \bar{s} < \delta/2$ \\
\textsc{Detect} $\rightarrow$ \textsc{Regulate} & confirmation: $s - s_o > s_c$ and $t - t_o \ge T_{\min}$; set $s_{\mathrm{ref}}$ from $k$ \\
\textsc{Regulate} $\rightarrow$ \textsc{Return} & release: operator keeps the finger open for $T_{\mathrm{open}}$ \\
\textsc{Return} $\rightarrow$ \textsc{Monitor} & re-arm: $q = q_h$ and spring back at baseline \\
\hline
\end{tabular}
\end{table}

\begin{figure*}[!t]
    \centering
    \includegraphics[width=\linewidth]{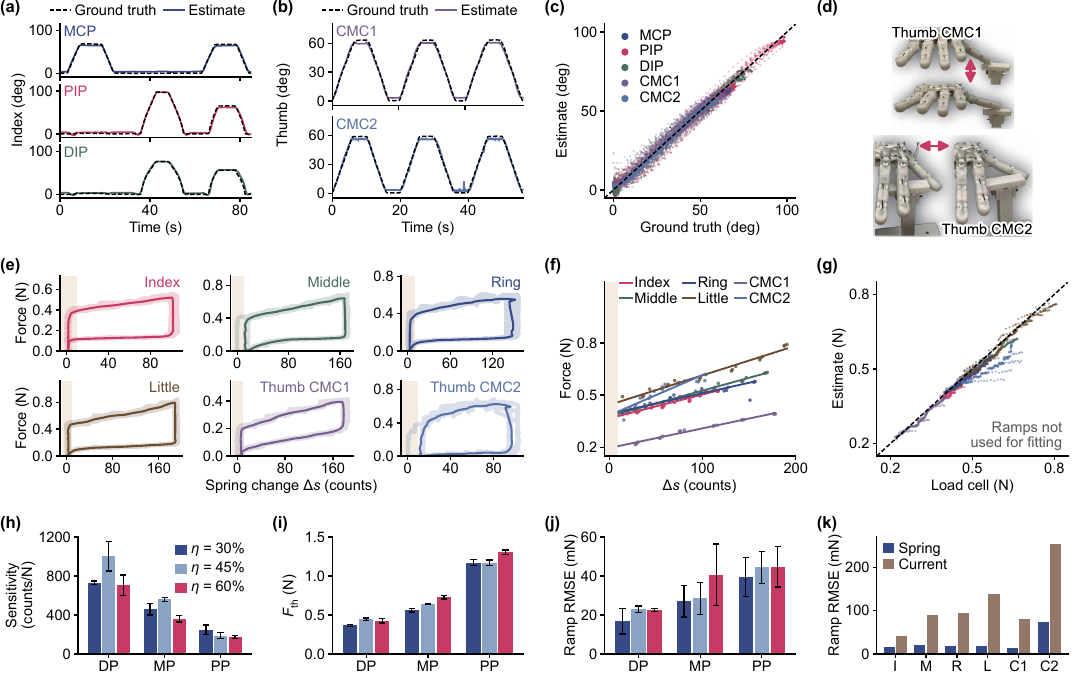}
    \caption{Self-sensing performance (black dashed lines indicate ground truth). (a)--(c) Joint angles in free motion: index finger, thumb CMC joints, and estimate versus ground truth. (d) Thumb loading directions. (e) Loading--unloading loops of the six channels for distal contact (bands: individual cycles; lines: mean loop; shaded: dead zone). (f) Force calibration during loading. (g) Estimate versus load cell on ramps (which are not used for fitting). (h)--(j) Sensitivity, activation threshold and ramp RMSE of the index finger for distal (DP), middle (MP) and proximal (PP) contact. (k) Ramp RMSE of the spring and of the motor-current baseline (I, M, R, L: index to little finger; C1, C2: thumb CMC).}
    \label{fig:selfSensingResults}
\end{figure*}

\paragraph{Contact detection}
\label{sec:contactDetection}
Contact is detected in each finger from the deviation $\Delta s = s - \bar{s}$ of the spring signal from an adaptive baseline $\bar{s}$. The baseline is initialized with the mean of the signal at start-up, whose standard deviation $\sigma_0$ characterizes the noise, and is then tracked by an exponential moving average, $\bar{s} \leftarrow (1-\alpha)\,\bar{s} + \alpha\, s$, so that slow drifts due to posture, tendon creep and preload are followed while the fast rise caused by a contact is not. A contact onset is flagged at $t_o$ when $\Delta s$ exceeds $\delta = \max(\kappa_\sigma \sigma_0, \delta_{\min})$. The contact is confirmed at $t_c$ once the spring has risen by a further $s_c$ counts since onset over at least $T_{\min}$\, (Fig.~\ref{fig:sharedControl}b); if $\Delta s$ falls below $\delta/2$ before, the onset is discarded as noise. Combining an amplitude and a persistence condition rejects impulsive noise and the brief spikes of fast free-space motion while still detecting sustained contact promptly.

\paragraph{Speed detection}
\label{sec:speedGraded}
Since the robot finger follows the operator's finger until contact, the operator's closing speed determines how fast the spring is loaded once the finger meets the object. The hand therefore measures this speed as the loading rate $k = \Delta S / \Delta T$, the slope of $\Delta s$ between onset and confirmation, evaluated over at most a window $T_w = 350$\,ms after onset to bound the decision latency (Fig.~\ref{fig:sharedControl}b). The speed is thus measured at the contact itself, on the robot side, without an additional sensor or tracking of the operator's finger velocity. Each contact is then assigned to one of two speed classes by a threshold $k^\ast$: slow if $k < k^\ast$ and fast otherwise.

\paragraph{Contact force regulation}
The speed class (slow or fast) sets the spring target of each regulated finger $i$ relative to its own baseline,
\begin{equation}
s_{\text{ref},i} = \bar{s}_i + m\,\bigl(\delta_i + s_c\bigr), \qquad
m = \begin{cases} 1, & k < k^\ast \\ m_{\text{fast}}, & k \ge k^\ast \end{cases}
\label{eq:target}
\end{equation}
so that a slow approach leads to a gentle grasp and a fast approach to a firm one, without any explicit command from the operator (Fig.~\ref{fig:sharedControl}c). The term $\delta_i + s_c$ is the spring change at which the contact was confirmed. A slow contact therefore holds the finger at this confirmation level, the lightest grasp that the detector can resolve, whereas a fast contact scales it by a factor $m_{\text{fast}} > 1$ to obtain a firmer grasp. Because $\delta_i$ and $s_c$ are defined in the counts of each channel, the same rule applies to every finger without a separate force calibration; the threshold $k^\ast$, whose value depends on the sensitivity of the channel, was set empirically. Once the contact is confirmed, the operator command for the finger is no longer executed, and its MCP motor is driven by a velocity command proportional to the spring error $e_i = s_{\text{ref},i} - s_i$,
\begin{equation}
\dot{q}_{\text{MCP},i} = \operatorname{sat}_{\dot q_{\max}}\!\left(K_p\, e_i\right), \qquad |e_i| > e_{\text{db}},
\label{eq:regulation}
\end{equation}
which drives the spring into the band $s_{\text{ref}} \pm e_{\text{db}}$, where the motor stops. The PIP--DIP motor follows through a fixed synergy, $\Delta q_{\text{PIP--DIP},i} = \rho\, \Delta q_{\text{MCP},i}$, measured from the posture at takeover. Because the operator command is not executed, further closing of the operator's hand does not compress the object further.

\paragraph{Per-finger state machine and allocation}
\label{sec:fsm}
\label{sec:authorityAllocation}
The three steps are combined in a state machine that runs independently for each of the six channels (Fig.~\ref{fig:sharedControl}a, Table~\ref{tab:fsm} and Algorithm~\ref{alg:sharedControl}). In \textsc{Monitor}, the operator has full control authority, $q = q_h$, and the baseline is tracked. An onset moves the finger to \textsc{Detect}, in which the operator still drives the finger while the contact and the loading rate are evaluated, and the confirmation moves it to \textsc{Regulate}. When the operator opens a regulated finger, the robot finger keeps its posture and the grasp is maintained; only if the opening persists for $T_{\text{open}}$ does the finger enter \textsc{Return}, in which the motor command is ramped to $q_h$ before the detector is re-armed. This delay distinguishes a deliberate release from tracking noise and brief unintended openings, e.g., when the ring and little fingers are occluded from the headset. In the normal cycle, a finger passes through the states \textsc{Monitor}, \textsc{Detect}, \textsc{Regulate} and \textsc{Return}.

Because the state machines run per finger, control authority can be allocated locally. A policy specifies which channels trigger a takeover and which fingers are then regulated, held or left to the operator. Two policies are used in this work. In the fragile-object and speed-graded tasks (Sections~\ref{sec:resFragile} and \ref{sec:resSpeedGraded}), a contact on any channel transfers the whole hand to the controller. The individual-finger allocation is used in the concurrent grasping and tool-use task (Section~\ref{sec:resToolUse}). Following the functional division of the human hand, in which the ulnar digits stabilize grasped objects while the thumb and index finger manipulate \cite{napier1956prehensile}, an ulnar policy then takes over only the middle, ring and little fingers, so that the hand holds one object while the operator manipulates another with the radial digits.

\begin{algorithm}[t]
\caption{Speed-Dictated Shared Control of One Finger}
\label{alg:sharedControl}
\begin{algorithmic}
\STATE \textbf{Input:} filtered spring signal $s(t)$, operator command $q_h(t)$
\STATE Initialize $\bar{s}$ and $\sigma_0$ over 2\,s; $\delta \leftarrow \max(\kappa_\sigma \sigma_0, \delta_{\min})$
\LOOP
    \IF{\textsc{Monitor}}
        \STATE $q \leftarrow q_h$; \ $\bar{s} \leftarrow (1-\alpha)\bar{s} + \alpha s$
        \STATE \textbf{if} $s - \bar{s} > \delta$ \textbf{then} $t_o \leftarrow t$, $s_o \leftarrow s$; go to \textsc{Detect}
    \ELSIF{\textsc{Detect}}
        \STATE $q \leftarrow q_h$
        \IF{$s - \bar{s} < \delta/2$}
            \STATE go to \textsc{Monitor} \hfill (noise)
        \ELSIF{$s - s_o > s_c$ \AND $t - t_o \ge T_{\min}$}
            \STATE $k \leftarrow$ loading rate over $[t_o, \min(t, t_o + T_w)]$
            \STATE $m \leftarrow 1$ if $k < k^\ast$, else $m_{\text{fast}}$
            \STATE $s_{\text{ref}} \leftarrow \bar{s} + m\,(\delta + s_c)$; go to \textsc{Regulate}
        \ENDIF
    \ELSIF{\textsc{Regulate}}
        \STATE $e \leftarrow s_{\text{ref}} - s$; \textbf{if} $|e| > e_{\text{db}}$ \textbf{then} $\dot q_{\text{MCP}} \leftarrow \operatorname{sat}(K_p e)$
        \STATE $\Delta q_{\text{PIP--DIP}} \leftarrow \rho\, \Delta q_{\text{MCP}}$
        \STATE \textbf{if} operator opens the finger for $T_{\text{open}}$ \textbf{then} go to \textsc{Return}
    \ELSIF{\textsc{Return}}
        \STATE ramp $q \rightarrow q_h$; \textbf{if} $q = q_h$ \textbf{then} re-arm; go to \textsc{Monitor}
    \ENDIF
\ENDLOOP
\end{algorithmic}
\end{algorithm}

\subsection{Experimental Setup}
\label{sec:setup}

\begin{figure}[!t]
    \centering
    \includegraphics[width=\linewidth]{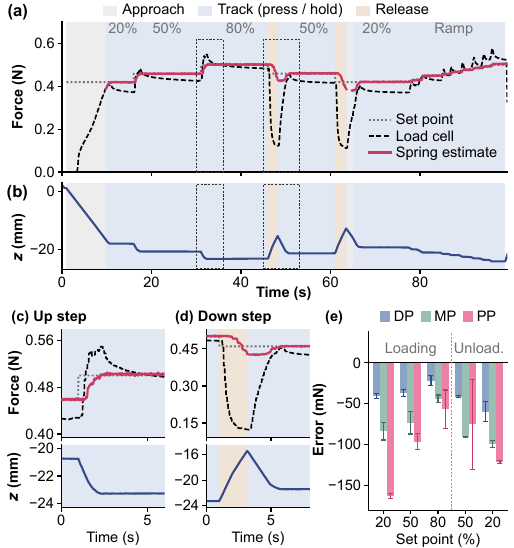}
    \caption{Closed-loop force control of the UR5 with the spring estimate as the only feedback. (a) Force and controller state. (b) Height of the UR5 tool. (c), (d) Enlarged up and down step. (e) Steady-state error (measured force minus set point) per step and contact location.}
    \label{fig:closedLoop}
\end{figure}

\begin{figure}[!t]
    \centering
    \includegraphics[width=\linewidth]{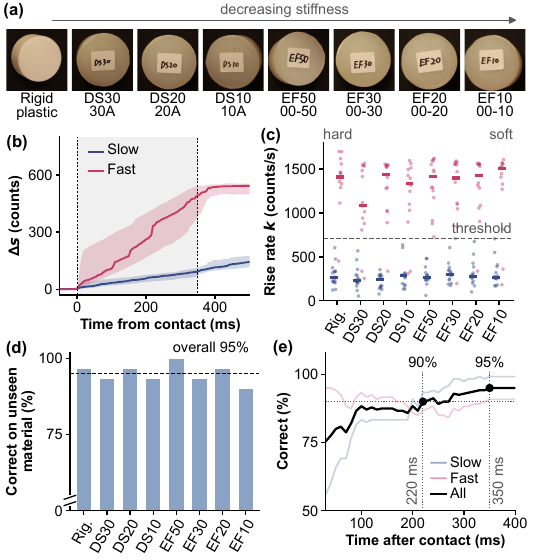}
    \caption{Classification of slow and fast approaches from the index-finger spring. (a) Contact samples of decreasing stiffness. (b) Spring change after contact (median and interquartile range). (c) Rise rate per sample and threshold. (d) Leave-one-sample-out accuracy per sample. (e) Accuracy versus the time after contact used for the decision.}
    \label{fig:speedIntent}
\end{figure}

\begin{figure}[!t]
    \centering
    \includegraphics[width=\linewidth]{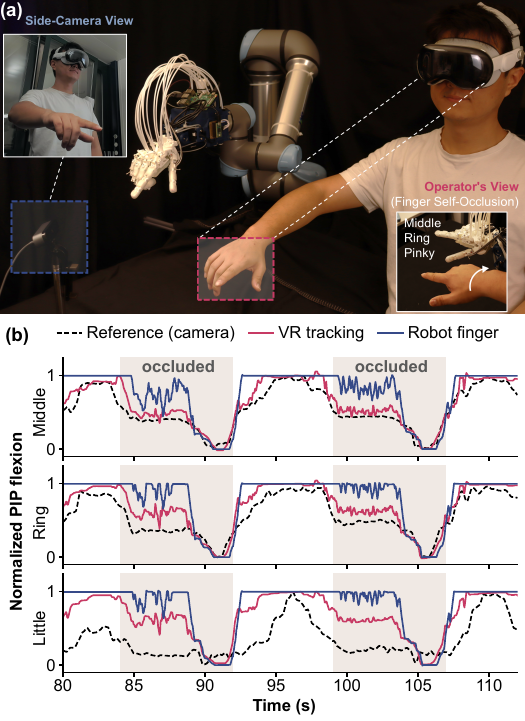}
    \caption{Self-occlusion of the ulnar fingers during VR teleoperation. (a) Experimental scene with the side camera. (b) Normalized PIP flexion of the middle, ring and little finger measured by the camera (reference), tracked by the headset, and executed by the robot; shaded: occlusion.}
    \label{fig:occlusion}
\end{figure}

\paragraph{Implementation}
The potentiometers (10\,k$\Omega$), one attached to each spring, are wired as voltage dividers and sampled by the analog-to-digital converter of a Teensy microcontroller, so that the spring signal is ratiometric and independent of the supply voltage. The AVP poses are streamed via gRPC to a host computer running ROS~2, which computes the arm and wrist commands, while the hand mapping and the low-level motor control run on a Raspberry Pi~4B on the hand.

\paragraph{Self-sensing calibration}
For joint-state estimation, the training data of each finger were three repetitions of a 90-s free-air sequence of MCP, PIP--DIP and combined flexion, with ground-truth angles from markers tracked in side-view video (Fig.~\ref{fig:selfSensingMethod}a,b). For force estimation, the hand was mounted on the UR5, which pressed a phalanx onto a two-axis load cell (HX711 amplifier) while the finger was held in a fixed posture; the spring and motor signals were logged at about 60\,Hz (Fig.~\ref{fig:selfSensingMethod}c). For the thumb, the load cell was placed so that the force loaded the sensed joint (Fig.~\ref{fig:selfSensingResults}d). Each run combined a staircase and a ramp mode (Fig.~\ref{fig:selfSensingMethod}d). In the staircase mode, the arm lowered the hand from the first contact in ten equal depth steps, held each for 2\,s and returned in the same steps, three times, with a maximum depth of 20, 15 and 10\,mm for the three phalanges (distal to proximal). In the ramp mode, it pressed to the same depth and returned at 1 or 4\,mm\,s$^{-1}$, twice per speed. Model \eqref{eq:linearForce} was fitted by least squares to the mean force and spring change over the second half of each loading dwell (30 points per run); the ramps were used only for validation. The posture is described by the normalized position $\eta$ of the MCP (or CMC) joint, from 0\% at full extension to 100\% at full flexion. All six channels were calibrated for distal contact at $\eta = 30\%$, and the index finger additionally for each phalanx at $\eta = 30\%$, 45\% and 60\%. The servos also report their current, which is available without any added hardware. To test whether the springs are needed, the same procedure was applied to the current of the servo that drives the sensed joint: a linear model $\hat{F} = c_0 + c_1 \Delta I$, where $\Delta I$ is the current change relative to the unloaded level, was fitted on the same static loading points and validated on the same ramps. To test the estimate as feedback (Fig.~\ref{fig:closedLoop}), the UR5 regulated the contact force by moving vertically with a proportional velocity command on the error between a set point and $\hat{F}$, with the finger held at $\eta = 30\%$ and the load cell used only as ground truth. The set point was stepped through 20, 50, 80, 50 and 20\% of the calibrated range (15\,s each), followed by a 20-s ramp; since the model describes only loading, lower set points were approached from below after a short retraction. Accuracy is reported as root-mean-square error (RMSE) and coefficient of determination ($R^2$).

\paragraph{Speed classification}
The index finger was closed by its MCP motor, without an operator, onto eight cylindrical samples of 20\,mm thickness: rigid plastic, Dragon Skin (Shore 30A, 20A, 10A) and Ecoflex (Shore 00-50, 00-30, 00-20, 00-10) (Fig.~\ref{fig:speedIntent}a). The closing speed was drawn at random from a slow (18--30\,deg\,s$^{-1}$) or a fast (90--120\,deg\,s$^{-1}$) range, and the UR5 positioned the fingertip above the sample so that the finger always moved freely before touching it. Each sample received 15 slow and 15 fast closures in random order (240 contacts); ten closures in air and seven slow closures on the rigid sample, distributed over the session, served to check for drift. Offline, contact was detected when the spring exceeded its pre-closure baseline by five standard deviations, and the rise rate $\bar{k}$, the offline counterpart of $k$, was computed as the mean increase of the spring over a window $T_w$ after contact. A single threshold on $\bar{k}$ was fitted on seven samples and tested on the eighth (leave-one-sample-out). This offline test assesses whether the loading rate separates slow from fast closures across object stiffness; the controller uses the same quantity over the shorter, contact-dependent interval between onset and confirmation, so that the offline accuracy is an upper bound for the online classification.

\paragraph{Teleoperation tasks}
The hand was teleoperated by two members of the research team, who became familiar with the setup in a short session of about 5\,min before the tasks. To quantify self-occlusion (Fig.~\ref{fig:occlusion}), a side camera with an unobstructed view recorded the operator's hand at 60\,fps while the fingers were flexed and extended, and reference PIP angles were extracted from this video with MediaPipe \cite{zhang2020mediapipe}, using the same angle definition as for the AVP skeleton. To evaluate whether the contact detection protects fragile objects (fragile-object task), the operator grasped a deformable plastic cup with MCP-dominant, full-finger and PIP--DIP-dominant flexion, with and without shared control; a confirmed contact on any channel held the posture of the whole hand. To evaluate whether the closing speed selects a suitable grasp (speed-graded task), a fragile plastic cup and a water bottle (220\,g) were each grasped and lifted 20 times in succession, the bottle additionally being rotated by 90\,deg. Each object was handled without shared control and with shared control in two conditions, in which the operator closed the hand slowly or quickly; the grasp level was not preset but selected online at every takeover. A trial was successful if the object was lifted without being crushed, dropped or slipping. In a continuous task, the operator then placed the cup with a slow approach and grasped the bottle to pour water into it. To evaluate the finger-level allocation (tool-use task), the ulnar policy was used while the operator manipulated a second object with the radial digits. How closely each digit followed the operator was quantified by a motion-following index (MFI), the fraction of human finger motion events (detected from the angular velocity, with a minimum displacement and duration) that the robot finger reproduced in the same direction within a time window,
\begin{equation}
\mathrm{MFI}_i = \frac{1}{N_i} \sum_{k=1}^{N_i} m_{i,k},
\label{eq:mfi}
\end{equation}
where $N_i$ is the number of human motion events of finger $i$ and $m_{i,k} \in \{0,1\}$ indicates whether event $k$ was matched.
\section{Experimental Results}
\label{sec:experiments}

\subsection{Self-Sensing Performance}
\label{sec:resSelfSensing}

We first tested whether the springs and motor positions alone suffice to estimate the state of the hand.

\begin{table}[t]
\centering
\caption{Self-Sensing Accuracy per Spring Channel}
\label{tab:selfSensingRMSE}
\renewcommand{\arraystretch}{1.15}\footnotesize\setlength{\tabcolsep}{3.2pt}
\begin{tabular}{l|cccccc}
\hline
 & Index & Middle & Ring & Little & CMC1 & CMC2 \\
\hline
\multicolumn{7}{@{}l}{\emph{Joint angle (joint loaded by the spring)}} \\
RMSE (deg)                & 3.7 & 3.4 & 3.6 & 3.7 & 2.3 & 2.5 \\
$R^2$                     & 0.98 & 0.98 & 0.98 & 0.98 & 0.99 & 0.99 \\
\hline
\multicolumn{7}{@{}l}{\emph{Fingertip force, spring \eqref{eq:linearForce}}} \\
Static $R^2$              & 0.93 & 0.99 & 0.94 & 0.97 & 0.99 & 0.84 \\
Static RMSE (mN)          & 12 & 8 & 13 & 17 & 6 & 27 \\
Ramp RMSE (mN)            & 17 & 21 & 19 & 19 & 15 & 74 \\
Ramp bias (mN)            & 14 & 18 & 16 & 10 & 12 & 69 \\
Hysteresis band (N)       & 0.32 & 0.39 & 0.36 & 0.47 & 0.17 & 0.45 \\
\hline
\multicolumn{7}{@{}l}{\emph{Fingertip force, motor-current baseline}} \\
Static $R^2$              & 0.14 & 0.01 & 0.12 & 0.56 & 0.03 & 0.53 \\
Ramp RMSE (mN)            & 42 & 90 & 96 & 139 & 81 & 255 \\
\hline
\multicolumn{7}{@{}p{0.95\linewidth}@{}}{\scriptsize Joint angle: five-fold blocked cross-validation. Force: distal contact at $\eta = 30\%$; ramps not used for fitting; bias = mean of measured minus estimated force.}
\end{tabular}
\end{table}

\begin{figure*}[!t]
    \centering
    \includegraphics[width=\linewidth]{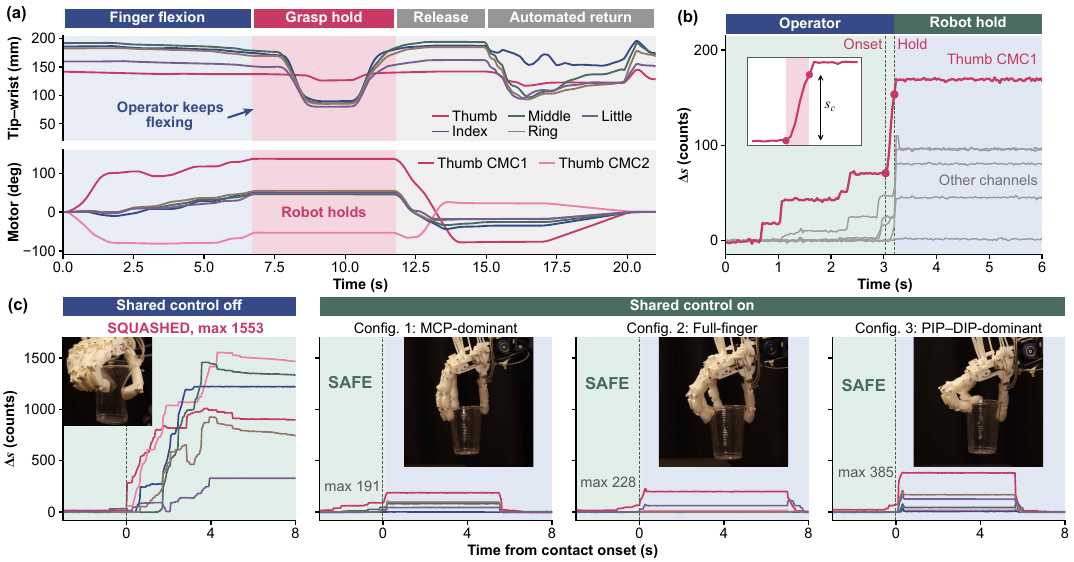}
    \caption{Teleoperated grasping of a fragile cup with the fragile-object policy. (a) Operator and robot motion during a grasp. (b) Contact detection on the thumb CMC1 spring and transfer of control authority to the hand. (c) Spring change with and without shared control for three grasp configurations.}
    \label{fig:res6_fragileCup}
\end{figure*}

\begin{figure*}[!t]
    \centering
    \includegraphics[width=\linewidth]{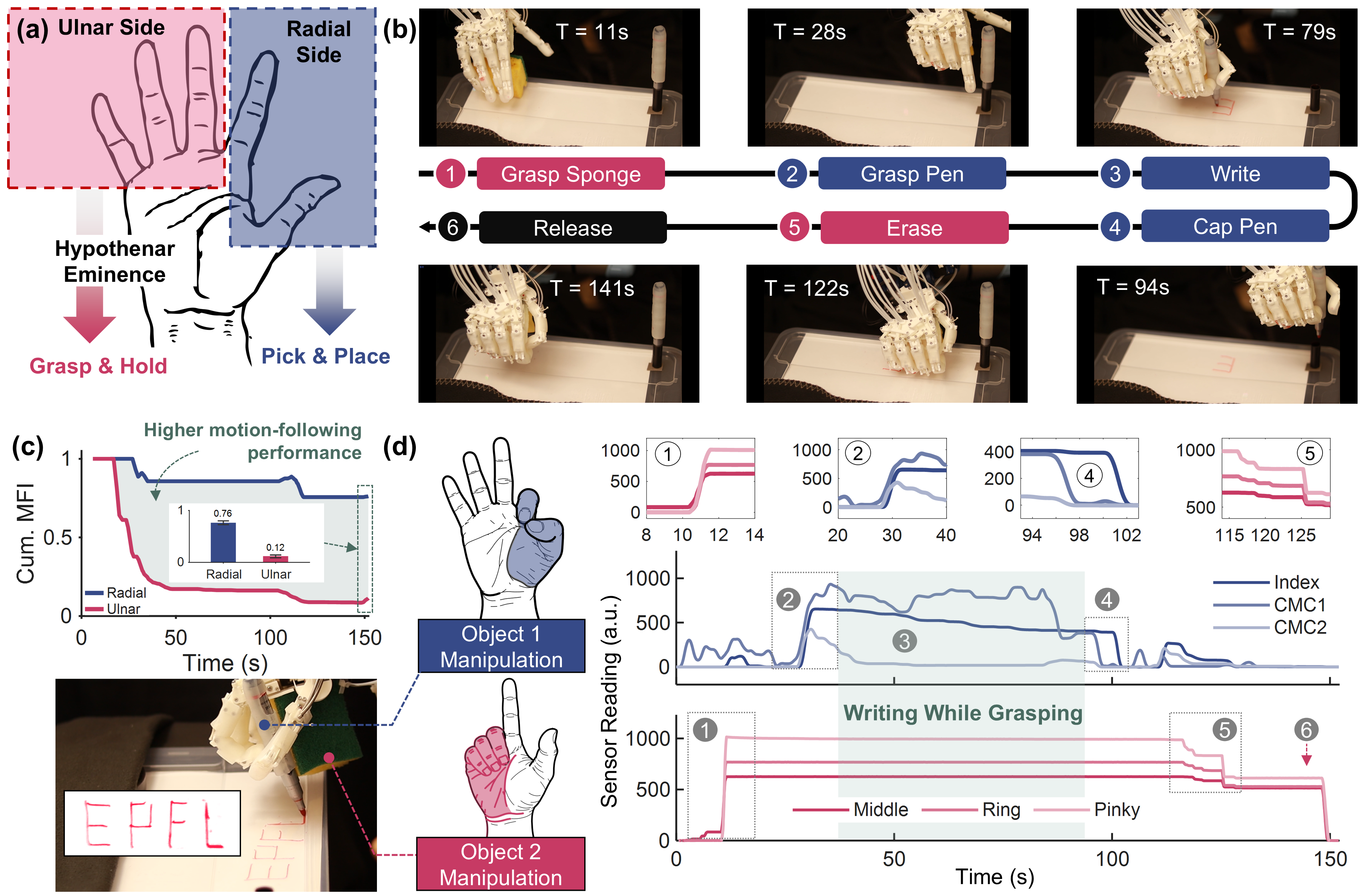}
    \caption{Finger-level allocation of control authority for concurrent grasping and tool use. (a) Functional division of the hand into ulnar (grasp-and-hold) and radial (precision manipulation) regions. (b) Sequence of sponge grasping, pen pick-up, writing, capping and erasing. (c) Motion-following index of the radial and ulnar digits. (d) Spring signals during distributed shared control; shading indicates writing.}
    \label{fig:res7_powerAssist}
\end{figure*}

\begin{figure*}[!t]
    \centering
    \includegraphics[width=1\linewidth]{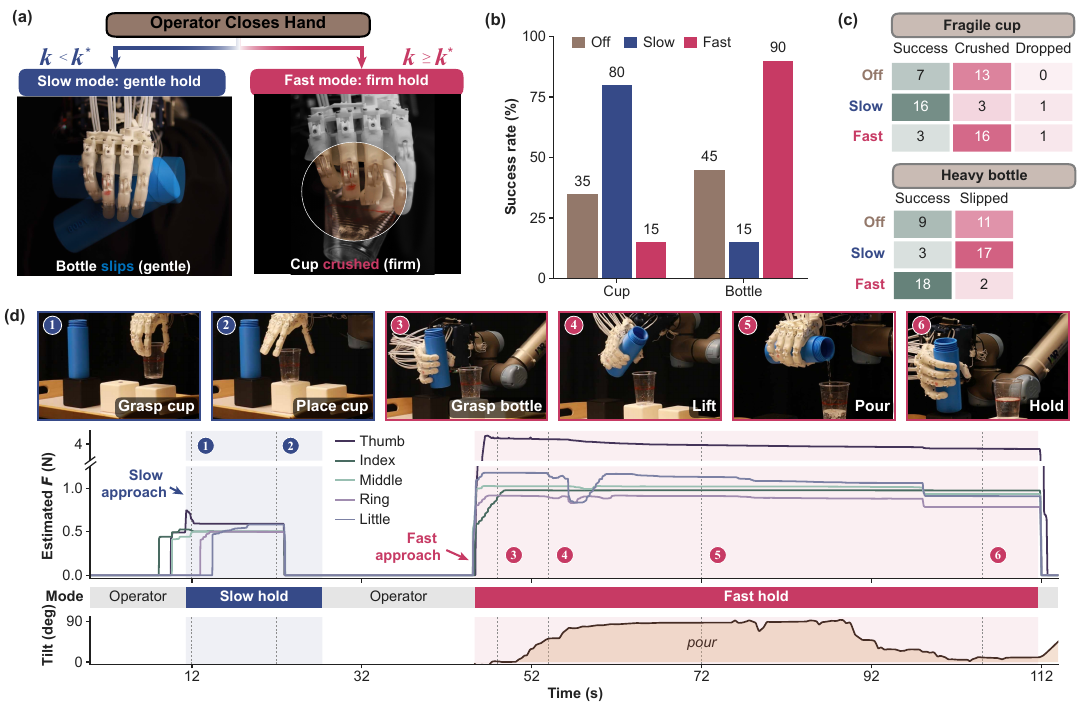}
    \caption{Speed-graded grasping. (a) The approach speed selects a gentle or a firm hold; a firm grasp crushes the fragile cup, and a gentle grasp lets the heavy bottle slip. (b) Success rates with shared control off and with shared control on in the slow and fast mode (20 trials each). (c) Outcomes of the same trials. (d) One cycle of the continuous task: video frames, estimated force on each finger, control mode and bottle tilt.}
    \label{fig:speedGraded}
\end{figure*}

\paragraph{Free-motion self-coupling}
Before estimating joint angles and contact forces, we quantified how much the spring signal changes without contact. Each actuated joint of the four fingers and the thumb was moved individually in free air. Bending the MCP joint of a finger or the CMC1 joint of the thumb raised the signal of the corresponding spring, mainly towards large flexion (Fig.~\ref{fig:selfSensingMethod}b), whereas motions of the PIP--DIP joints and of the other joints left the springs practically unchanged, apart from a coupling from CMC1 to CMC2. The self-coupling is therefore confined to the joint that loads the spring and is repeatable. It carries information on the posture that the joint-angle model exploits, and it justifies the posture-dependent term $s_0(q)$ in \eqref{eq:springDecomposition} and the adaptive baseline used for contact detection.

\paragraph{Joint angles}
Using only the motor positions and the spring signal, the SVR reproduced the joint angles of the fingers and the thumb through flexion, holding and extension (Fig.~\ref{fig:selfSensingResults}a--c; Video~S2). In five-fold blocked cross-validation, in which each contiguous fifth of a recording was predicted by a model trained on the other four, the RMSE was about 3\,deg on average (at most 3.7\,deg for any joint), with $R^2 \geq 0.98$ for every joint (Table~\ref{tab:selfSensingRMSE}). The points of all joints lay close to the identity line over the full range of motion (Fig.~\ref{fig:selfSensingResults}c), and the PIP and DIP joints, which carry no sensor, were estimated as accurately as the MCP joint.

\paragraph{Hysteresis of the tendon--spring transmission}
Over complete loading--unloading cycles, all six channels showed the same loop shape (Fig.~\ref{fig:selfSensingResults}e). On loading, i.e., while the force increased, the force first rose by about 0.4\,N on average while the spring remained in the dead zone, after which force and spring change increased together approximately linearly. On unloading, the force dropped at an almost constant spring change and then decreased along a lower curve until the spring returned to its baseline. For five channels, the individual cycles lay close to the mean loop, including the ramps at 4\,mm\,s$^{-1}$, so that the loop was repeatable and largely independent of the loading speed; the cycles of the thumb CMC2 channel, which is loaded sideways, scattered more. The force difference between loading and unloading at the same spring change was 0.36\,N on average, with the narrowest loop for the thumb CMC1 channel (Table~\ref{tab:selfSensingRMSE}). The force model was therefore restricted to loading, which is the phase relevant for detecting and regulating contact.

\paragraph{Fingertip force}
On the static loading steps, the activation threshold $F_{\text{th}}$ of \eqref{eq:linearForce} was 0.36\,N on average. During loading, the force was well described by the linear model for five of the six channels (mean $R^2 = 0.96$, mean RMSE 11\,mN; Fig.~\ref{fig:selfSensingResults}f). On the continuous ramps, which were not used for fitting, the estimate of these five channels followed the load cell up to about 0.8\,N with a mean RMSE of 18\,mN (Fig.~\ref{fig:selfSensingResults}g), most of which was a consistent underestimate of 14\,mN on average. The estimate of the thumb CMC2 channel fell increasingly below the measured force as the load increased, in line with its lower static fit and its sideways loading.

\paragraph{Motor-current baseline}
To demonstrate that the series spring is needed, we compared it with the servo current, which is available without any added hardware. Fitted with the same procedure, the current failed to describe the fingertip force well (Fig.~\ref{fig:selfSensingResults}k). On the static loading points, the current explained on average 23\% of the force variance, compared with 94\% for the spring. On the ramps, the current-based estimate reached a mean RMSE of 90\,mN for the four fingers and CMC1 and 255\,mN for CMC2, larger than the error of the spring for every channel. In position control, the servo current reflects the position error of the servo controller and the friction in the gearbox and Bowden tubes rather than the tendon tension, and the change caused by sub-newton fingertip forces was comparable to the current noise at rest (standard deviation up to 1.6\,mA).

\paragraph{Phalangeal Contact}
The spring also responded to contacts on the middle and proximal phalanges of the index finger (Fig.~\ref{fig:selfSensingResults}h--j). The force remained linear in the spring change during loading (mean $R^2 = 0.96$), but from the distal to the proximal phalanx the sensitivity decreased and the activation threshold rose from about 0.4\,N to 1.2\,N on average, as expected from the shorter lever arm. The posture changed the threshold by at most 0.2\,N. With location-specific models, the mean ramp RMSE was 21, 32 and 43\,mN for distal, middle and proximal contact. The spring detects contacts along the whole finger, but a quantitative force estimate requires knowing which phalanx is loaded.

\paragraph{Force estimate as feedback}
\label{sec:resClosedLoop}
We checked that the estimate is sufficient to close a force loop on its own (Fig.~\ref{fig:closedLoop}). With $\hat{F}$ as the only feedback, the UR5 settled within 2.6\,s on average for up steps, with an overshoot of at most 50\,mN, and within 5.3\,s for down steps, and it tracked the ramp with an RMSE of about 20\,mN. For distal contact, the measured force stayed on average 40\,mN below the set point, although the estimate settled within 2\,mN of it, and this deviation grew towards proximal contact (40, 80 and 100\,mN for the distal, middle and proximal phalanx of the index finger; Fig.~\ref{fig:closedLoop}e), in line with the phalanx-dependent calibration.

\subsection{Speed Detection at Contact}
\label{sec:resContactDetection}

For the closing speed to serve as a force command, the loading rate at contact is supposed to reflect the speed of the finger rather than the stiffness of the object, so that a single threshold applies to unknown objects. We therefore closed the index finger at a slow or a fast speed onto eight samples ranging from rigid plastic to the softest Ecoflex (Fig.~\ref{fig:speedIntent}a; Video~S3). Over $T_w = 350$\,ms after contact, the median rise rate was about 270\,counts\,s$^{-1}$ for slow and 1400\,counts\,s$^{-1}$ for fast closures, and the two groups remained separated on every sample (Fig.~\ref{fig:speedIntent}b,c). With leave-one-sample-out validation, a single threshold classified 95\% of the 240 contacts correctly, and at least 90\% on every sample (Fig.~\ref{fig:speedIntent}d). Eleven of the twelve errors were fast closures classified as slow, i.e., in the conservative direction that yields a gentle instead of a firm grasp. The threshold varied by less than 3\% across folds (about 700\,counts\,s$^{-1}$), and the control closures showed no drift over the session. The accuracy exceeded 90\% from $T_w = 220$\,ms (Fig.~\ref{fig:speedIntent}e).

\subsection{Contact-Triggered Takeover of Fragile Objects}
\label{sec:resFragile}

This experiment tested whether the contact detection alone transfers control authority early enough to prevent over-compression, regardless of how the operator closes the hand. In the grasp of Fig.~\ref{fig:res6_fragileCup}a, closed mainly at the MCP joints, the hand followed the operator until the thumb touched the cup at $t = 6.7$\,s. The contact was confirmed on the thumb CMC1 channel within 0.17\,s, and control authority passed to the hand (Fig.~\ref{fig:res6_fragileCup}b). The operator kept closing the hand, but the robot did not execute this further flexion, which would otherwise have compressed the cup. Before contact, the CMC1 spring rose in small steps of less than 30\,counts as the thumb was flexed, which, owing to the adaptive baseline, did not trigger the takeover. For all three grasp configurations (Fig.~\ref{fig:res6_fragileCup}c), the springs kept rising up to about 1550\,counts without shared control and the cup was squashed, whereas with shared control the hand held its posture immediately after contact and the cup remained intact (Video~S4).

\subsection{Finger-Level Allocation of Control Authority}
\label{sec:resToolUse}

\paragraph{Self-occlusion of the ulnar fingers}
\label{sec:resOcclusion}
We first examined why the ulnar fingers benefit from being held by the hand. During teleoperation, the middle, ring and little fingers were frequently occluded from the headset cameras by the index finger and the back of the hand (Fig.~\ref{fig:occlusion}a). Outside occlusion, the headset tracking followed the side-camera reference, and the robot reproduced the operator's motion. When the ring and little fingers were hidden during extension, however, the headset kept them flexed, reporting a PIP angle above 40\,deg whereas the reference measured below 35\,deg (Fig.~\ref{fig:occlusion}b, shaded), and the robot executed this error, with its PIP--DIP motors above 125\,deg instead of below 50\,deg in a correctly tracked cycle (Video~S5). The middle finger was less affected.

\paragraph{Concurrent grasping and tool use}
We then tested whether the per-finger allocation lets the hand hold one object while the operator manipulates another. With the ulnar policy, the ulnar fingers grasped a sponge, entered \textsc{Regulate} and held it, while the thumb and index finger remained under teleoperation, picked up a pen, wrote on a board and capped the pen (Fig.~\ref{fig:res7_powerAssist}b; Video~S6). The retained sponge was then used to erase the writing, so that the whole sequence was completed without releasing or regrasping the sponge. The index finger, under teleoperation, followed the operator with an MFI \eqref{eq:mfi} of 0.76, whereas the held middle finger reached only 0.12 (Fig.~\ref{fig:res7_powerAssist}c). During writing, the pen contact loaded the radial springs while their motors kept tracking the operator, and during erasing the ulnar spring signals decreased as the sponge was pressed against the board and the hypothenar region of the hand (Fig.~\ref{fig:res7_powerAssist}d). Throughout the task, the two functional regions of the hand thus remained under distinct control authority.

\subsection{Speed-Dictated Grasping}
\label{sec:resSpeedGraded}

Finally, we tested whether the operator can use the closing speed in teleoperation to obtain the grasp that each object requires, using two objects with opposite requirements: a fragile plastic cup, which is crushed by a firm grasp, and a water bottle, which slips from a gentle one (Fig.~\ref{fig:speedGraded}a; Video~S7).

\paragraph{Shared control off versus on}
With the speed suited to the object, shared control improved both tasks compared with direct teleoperation (Fig.~\ref{fig:speedGraded}b): slow closing increased the success rate for the cup from 35\% to 80\%, and fast closing that for the bottle from 45\% to 90\%. Without shared control, 13 of the 20 cup grasps crushed the cup and 11 of the 20 bottle grasps let the bottle slip (Fig.~\ref{fig:speedGraded}c), i.e., the operator could not reliably apply a suitable force without haptic feedback. With the unsuited speed, the same failures dominated (Fig.~\ref{fig:speedGraded}c).

\paragraph{Switching within one task}
In the continuous task, the operator performed the water-pour task by selecting different closing speeds in a single trial (Fig.~\ref{fig:speedGraded}d). The slow closure on the cup resulted in a gentle hold with estimated finger forces around 0.5\,N, which ended when the cup had been placed. The subsequent fast closure on the bottle resulted in a firm hold with about 1\,N on the fingers and a higher value on the thumb, which was maintained while the bottle was tilted by about 90\,deg to pour and returned upright. Note that these forces were estimated with the distal-contact models instead of measured grip forces, which illustrates the effectiveness of the speed-dictated switching between a gentle and a firm grasp within one task without any explicit command.
\section{Discussion and Conclusion}
\label{sec:discussion}

This paper demonstrates that the series springs of a compliant tendon-driven hand allow a teleoperator to dictate the grasp force through the closing speed. Read by potentiometers in the forearm-mounted motor box, the springs enable the hand to detect contacts and measure the loading rate at contact, which reflects the closing speed rather than the stiffness of the object. On this basis, the hand takes over the grasp at contact and selects a gentle or a firm grasp, which improves the success of both delicate and power grasps compared with direct teleoperation and enables the operator to switch between them within one task without any explicit command. In addition, these same springs can estimate the joint angles and fingertip forces without any sensors on the fingers, whereas the servo current (motor-only approach) shows compromised performance. In our framework, the per-finger state machines also enable the hand to hold one object while the operator manipulates another object simultaneously. Here, the elastic elements in our compliant hands are dimensioned not only for their mechanical role but also for their sensing role.

Several limitations remain. First, the speed grading currently distinguishes two force levels and its thresholds were mainly tuned empirically. Second, the spring stiffness trades sensitivity against range, so that the force estimate saturates for heavy loads. Finally, with no external tactile sensors equipped, the sensitivity depends on the loaded phalanx, and a single spring per finger struggles to localize the contact or detect tangential forces or slip. Future work could focus on grading the grasp force continuously with the closing speed, detecting slip from the spring signal, feeding the estimated forces back to the operator, and evaluating the framework with multiple operators. More broadly, these results indicate that mechanical compliance in soft and compliant hands is not only a means of passive adaptation but also an intrinsic sensing resource for physical human--robot collaboration.

\section*{ACKNOWLEDGMENT}
This work was supported by Innosuisse, the Swiss Innovation Agency, under Grant 119.806 IP-ENG.

\balance
\bibliographystyle{IEEEtran}
\bibliography{main}

\begin{thebibliography}{10}
\providecommand{\url}[1]{#1}
\csname url@rmstyle\endcsname
\providecommand{\newblock}{\relax}
\providecommand{\bibinfo}[2]{#2}
\providecommand\BIBentrySTDinterwordspacing{\spaceskip=0pt\relax}
\providecommand\BIBentryALTinterwordstretchfactor{4}
\providecommand\BIBentryALTinterwordspacing{\spaceskip=\fontdimen2\font plus
\BIBentryALTinterwordstretchfactor\fontdimen3\font minus
  \fontdimen4\font\relax}
\providecommand\BIBforeignlanguage[2]{{%
\expandafter\ifx\csname l@#1\endcsname\relax
\typeout{** WARNING: IEEEtran.bst: No hyphenation pattern has been}%
\typeout{** loaded for the language `#1'. Using the pattern for}%
\typeout{** the default language instead.}%
\else
\language=\csname l@#1\endcsname
\fi
#2}}

\bibitem{zhang2018deep}
T.~Zhang, Z.~McCarthy, O.~Jow, D.~Lee, X.~Chen, K.~Goldberg, and P.~Abbeel,
  ``Deep imitation learning for complex manipulation tasks from virtual reality
  teleoperation,'' in \emph{2018 IEEE international conference on robotics and
  automation (ICRA)}.\hskip 1em plus 0.5em minus 0.4em\relax Ieee, 2018, pp.
  5628--5635.

\bibitem{qin2023anyteleop}
Y.~Qin, W.~Yang, B.~Huang, K.~Van~Wyk, H.~Su, X.~Wang, Y.-W. Chao, and D.~Fox,
  ``Anyteleop: A general vision-based dexterous robot arm-hand teleoperation
  system,'' \emph{arXiv preprint arXiv:2307.04577}, 2023.

\bibitem{handa2020dexpilot}
A.~Handa, K.~Van~Wyk, W.~Yang, J.~Liang, Y.-W. Chao, Q.~Wan, S.~Birchfield,
  N.~Ratliff, and D.~Fox, ``Dexpilot: Vision-based teleoperation of dexterous
  robotic hand-arm system,'' in \emph{2020 IEEE International Conference on
  Robotics and Automation (ICRA)}.\hskip 1em plus 0.5em minus 0.4em\relax IEEE,
  2020, pp. 9164--9170.

\bibitem{gao2022efficient}
Q.~Gao, Z.~Ju, Y.~Chen, Q.~Wang, and C.~Chi, ``An efficient rgb-d hand gesture
  detection framework for dexterous robot hand-arm teleoperation system,''
  \emph{IEEE Transactions on Human-Machine Systems}, vol.~53, no.~1, pp.
  13--23, 2022.

\bibitem{shaw2024bimanual}
K.~Shaw, Y.~Li, J.~Yang, M.~K. Srirama, R.~Liu, H.~Xiong, R.~Mendonca, and
  D.~Pathak, ``Bimanual dexterity for complex tasks,'' \emph{arXiv preprint
  arXiv:2411.13677}, 2024.

\bibitem{cheng2024open}
X.~Cheng, J.~Li, S.~Yang, G.~Yang, and X.~Wang, ``Open-television:
  Teleoperation with immersive active visual feedback,'' \emph{arXiv preprint
  arXiv:2407.01512}, 2024.

\bibitem{ding2024bunny}
R.~Ding, Y.~Qin, J.~Zhu, C.~Jia, S.~Yang, R.~Yang, X.~Qi, and X.~Wang,
  ``Bunny-visionpro: Real-time bimanual dexterous teleoperation for imitation
  learning,'' \emph{arXiv preprint arXiv:2407.03162}, 2024.

\bibitem{junge2025adapt}
K.~Junge and J.~Hughes, ``Adapt-teleop: robotic hand with human matched
  embodiment enables dexterous teleoperated manipulation,'' \emph{npj
  Robotics}, vol.~3, no.~1, p.~31, 2025.

\bibitem{li2020mobile}
S.~Li, J.~Jiang, P.~Ruppel, H.~Liang, X.~Ma, N.~Hendrich, F.~Sun, and J.~Zhang,
  ``A mobile robot hand-arm teleoperation system by vision and imu,'' in
  \emph{2020 IEEE/RSJ International Conference on Intelligent Robots and
  Systems (IROS)}.\hskip 1em plus 0.5em minus 0.4em\relax IEEE, 2020, pp.
  10\,900--10\,906.

\bibitem{li2021survey}
R.~Li, H.~Wang, and Z.~Liu, ``Survey on mapping human hand motion to robotic
  hands for teleoperation,'' \emph{IEEE Transactions on Circuits and Systems
  for Video Technology}, vol.~32, no.~5, pp. 2647--2665, 2021.

\bibitem{li2023classification}
G.~Li, Q.~Li, C.~Yang, Y.~Su, Z.~Yuan, and X.~Wu, ``The classification and new
  trends of shared control strategies in telerobotic systems: A survey,''
  \emph{IEEE Transactions on Haptics}, vol.~16, no.~2, pp. 118--133, 2023.

\bibitem{huang2025human}
Y.~Huang, D.~Fan, D.~Yan, W.~Qi, G.~Deng, Z.~Shao, Y.~Luo, D.~Li, Z.~Wang,
  Q.~Liu, \emph{et~al.}, ``Human-robot collaborative tele-grasping in clutter
  with five-fingered robotic hands,'' \emph{IEEE Robotics and Automation
  Letters}, 2025.

\bibitem{michelman1994shared}
P.~Michelman and P.~Allen, ``Shared autonomy in a robot hand teleoperation
  system,'' in \emph{Proceedings of IEEE/RSJ International Conference on
  Intelligent Robots and Systems (IROS'94)}, vol.~1.\hskip 1em plus 0.5em minus
  0.4em\relax IEEE, 1994, pp. 253--259.

\bibitem{griffin2003feedback}
W.~B. Griffin, W.~R. Provancher, and M.~R. Cutkosky, ``Feedback strategies for
  shared control in dexterous telemanipulation,'' in \emph{Proceedings 2003
  IEEE/RSJ International Conference on Intelligent Robots and Systems (IROS
  2003)(Cat. No. 03CH37453)}, vol.~3.\hskip 1em plus 0.5em minus 0.4em\relax
  IEEE, 2003, pp. 2791--2796.

\bibitem{zhu2023shared}
Y.~Zhu, B.~Jiang, Q.~Chen, T.~Aoyama, and Y.~Hasegawa, ``A shared control
  framework for enhanced grasping performance in teleoperation,'' \emph{IEEE
  Access}, vol.~11, pp. 69\,204--69\,215, 2023.

\bibitem{laghi2022target}
M.~Laghi, L.~Raiano, F.~Amadio, F.~Rollo, A.~Zunino, and A.~Ajoudani, ``A
  target-guided telemanipulation architecture for assisted grasping,''
  \emph{IEEE Robotics and Automation Letters}, vol.~7, no.~4, pp. 8759--8766,
  2022.

\bibitem{kim2020six}
U.~Kim, H.~Jeong, H.~Do, J.~Park, and C.~Park, ``Six-axis force/torque
  fingertip sensor for an anthropomorphic robot hand,'' \emph{IEEE Robotics and
  Automation Letters}, vol.~5, no.~4, pp. 5566--5572, 2020.

\bibitem{mao2025ultralight}
Q.~Mao, Z.~Liao, S.~Liu, J.~Yuan, and R.~Zhu, ``An ultralight, tiny, flexible
  six-axis force/torque sensor enables dexterous fingertip manipulations,''
  \emph{Nature Communications}, vol.~16, no.~1, p. 5693, 2025.

\bibitem{xu2024soft}
Y.~Xu, S.~Zhang, S.~Li, Z.~Wu, Y.~Li, Z.~Li, X.~Chen, C.~Shi, P.~Chen,
  P.~Zhang, \emph{et~al.}, ``A soft magnetoelectric finger for robots’
  multidirectional tactile perception in non-visual recognition environments,''
  \emph{npj Flexible Electronics}, vol.~8, no.~1, p.~2, 2024.

\bibitem{li2024tactile}
H.~Li, C.~J. Ford, C.~Lu, Y.~Lin, M.~Bianchi, M.~G. Catalano, E.~Psomopoulou,
  and N.~F. Lepora, ``Tactile softhand-a: 3d-printed, tactile, highly
  underactuated, anthropomorphic robot hand with an antagonistic tendon
  mechanism,'' \emph{The International Journal of Robotics Research}, p.
  02783649251379516, 2024.

\bibitem{almeida2025role}
J.~D. Almeida, E.~Falotico, C.~Laschi, and J.~Santos-Victor, ``The role of
  touch: Towards optimal tactile sensing distribution in anthropomorphic hands
  for dexterous in-hand manipulation,'' \emph{arXiv preprint arXiv:2509.14984},
  2025.

\bibitem{zhu2025forces}
Y.~Zhu, M.~Hao, X.~Zhu, Q.~Bateux, A.~Wong, and A.~M. Dollar, ``Forces for
  free: Vision-based contact force estimation with a compliant hand,''
  \emph{Science Robotics}, vol.~10, no. 103, p. eadq5046, 2025.

\bibitem{della2018toward}
C.~Della~Santina, C.~Piazza, G.~Grioli, M.~G. Catalano, and A.~Bicchi, ``Toward
  dexterous manipulation with augmented adaptive synergies: The pisa/iit
  softhand 2,'' \emph{IEEE Transactions on Robotics}, vol.~34, no.~5, pp.
  1141--1156, 2018.

\bibitem{junge2025spatially}
K.~Junge and J.~Hughes, ``Spatially distributed biomimetic compliance enables
  robust anthropomorphic robotic manipulation,'' \emph{Communications
  Engineering}, vol.~4, no.~1, p.~76, 2025.

\bibitem{jami1992golgi}
L.~Jami, ``Golgi tendon organs in mammalian skeletal muscle: functional
  properties and central actions,'' \emph{Physiological Reviews}, vol.~72,
  no.~3, pp. 623--666, 1992.

\bibitem{kim2023multiparameter}
S.-H. Kim, U.~Jeong, and K.-J. Cho, ``Multiparameter remote contact force
  sensor with embedded bend sensing for tendon-driven hand robots,''
  \emph{IEEE/ASME Transactions on Mechatronics}, vol.~29, no.~1, pp. 557--566,
  2023.

\bibitem{yi2023underactuated}
J.~Yi, B.~Kim, K.-J. Cho, and Y.-L. Park, ``Underactuated robotic gripper with
  fiber-optic force sensing tendons,'' \emph{IEEE Robotics and Automation
  Letters}, vol.~8, no.~11, pp. 7607--7614, 2023.

\bibitem{lee2025tendon}
J.-H. Lee, J.~Park, and K.-J. Cho, ``Tendon-based proprioception in an
  anthropomorphic underactuated robotic hand with series elastic actuators,''
  \emph{arXiv preprint arXiv:2509.12969}, 2025.

\bibitem{bridgwater2012robonaut}
L.~B. Bridgwater, C.~A. Ihrke, M.~A. Diftler, M.~E. Abdallah, N.~A. Radford,
  J.~M. Rogers, S.~Yayathi, R.~S. Askew, and D.~M. Linn, ``The {Robonaut} 2
  hand -- designed to do work with tools,'' in \emph{Proc. IEEE Int. Conf.
  Robot. Autom. (ICRA)}, 2012.

\bibitem{yi2026multifingered}
J.~Yi, W.~J. Chung, J.~Lee, H.~Muzammal, J.~Park, Y.~S. Park, and Y.-L. Park,
  ``A multifingered robotic hand with fiber-optic force and tactile sensing for
  remote manipulation,'' \emph{IEEE Transactions on Robotics}, vol.~42, 2026.

\bibitem{marteniuk1987constraints}
R.~G. Marteniuk, C.~L. MacKenzie, M.~Jeannerod, S.~Athenes, and C.~Dugas,
  ``Constraints on human arm movement trajectories,'' \emph{Canadian Journal of
  Psychology}, vol.~41, no.~3, pp. 365--378, 1987.

\bibitem{zhang2026modular}
Y.~Zhang, H.~Wang, and J.~Hughes, ``Modular anthropomorphic hand design via
  multi-parameter finger benchmarking and selection,'' \emph{arXiv preprint
  arXiv:2606.11826}, 2026.

\bibitem{park2024avp}
Y.~Park and P.~Agrawal, ``Using apple vision pro to train and control robots,''
  \url{https://github.com/Improbable-AI/VisionProTeleop}, 2024.

\bibitem{napier1956prehensile}
J.~R. Napier, ``The prehensile movements of the human hand,'' \emph{The Journal
  of Bone and Joint Surgery. British Volume}, vol. 38-B, no.~4, pp. 902--913,
  1956.

\bibitem{zhang2020mediapipe}
F.~Zhang, V.~Bazarevsky, A.~Vakunov, A.~Tkachenka, G.~Sung, C.-L. Chang, and
  M.~Grundmann, ``{MediaPipe} hands: On-device real-time hand tracking,''
  \emph{arXiv preprint arXiv:2006.10214}, 2020.

\end{thebibliography}

\end{document}